\documentclass{article}

\usepackage{microtype}
\usepackage{graphicx}
\usepackage{subcaption}
\usepackage{booktabs} %

\usepackage{hyperref}

\usepackage[accepted]{icml2026}

\usepackage{amsmath}
\usepackage{amssymb}
\usepackage{mathtools}
\usepackage{amsthm}
\usepackage{hyperref}       %
\usepackage{url}            %
\usepackage{booktabs}       %
\usepackage{amsfonts}       %
\usepackage{nicefrac}       %
\usepackage{microtype}      %
\usepackage{xcolor}         %
\usepackage{algorithm}
\usepackage{algorithmic} %

\usepackage[capitalize,noabbrev]{cleveref}

\theoremstyle{plain}
\newtheorem{theorem}{Theorem}[section]
\newtheorem{proposition}[theorem]{Proposition}
\newtheorem{lemma}[theorem]{Lemma}
\newtheorem{corollary}[theorem]{Corollary}
\theoremstyle{definition}

\newtheorem{assumption}[theorem]{Assumption}
\newtheorem{example}{Example}
\theoremstyle{remark}

\usepackage[textsize=tiny]{todonotes}

\icmltitlerunning{On Optimization Complexity of Second-Order Certified Unlearning}

\newcommand{\vzero}{\boldsymbol{0}}

\newcommand{\R}{\mathbb{R}}

\renewcommand{\P}{\mathbb{P}}
\newcommand{\E}{\mathbb{E}}

\newcommand{\U}{\mathbb{U}}
\newcommand{\argmin}{\mathop{\mathrm{arg\,min}}}

\newcommand{\dom}{\operatorname{dom}}

\renewcommand{\aa}{\boldsymbol{a}}

\let\ggg\gg
\renewcommand{\gg}{\boldsymbol{g}}

\let\lll\ll
\renewcommand{\ll}{\boldsymbol{l}}

\renewcommand{\ss}{\boldsymbol{s}}

\providecommand{\vv}{\boldsymbol{v}}

\providecommand{\xx}{\boldsymbol{x}}
\providecommand{\yy}{\boldsymbol{y}}
\providecommand{\zz}{\boldsymbol{z}}

\renewcommand{\AA}{\boldsymbol{A}}
\newcommand{\BB}{\boldsymbol{B}}

\newcommand{\HH}{\boldsymbol{H}}
\newcommand{\II}{\boldsymbol{I}}
\newcommand{\xixi}{\boldsymbol{\xi}}

\newcommand{\beq}{\begin{equation}}
	\newcommand{\eeq}{\end{equation}}
\newcommand{\ba}{\begin{array}}
	\newcommand{\ea}{\end{array}}
\newcommand{\la}{\langle}
\newcommand{\ra}{\rangle}

\begin{document}

\twocolumn[
  \icmltitle{On Optimization Complexity of Second-Order Certified Unlearning}

  \icmlsetsymbol{equal}{*}

  \begin{icmlauthorlist}
    \icmlauthor{Nikita Doikov}{cornell}
    \icmlauthor{Anastasia Koloskova}{uzh}
  \end{icmlauthorlist}

  \icmlaffiliation{cornell}{Cornell University, Ithaca, USA}
  \icmlaffiliation{uzh}{University of Zurich, Zurich, Switzerland}

  \icmlcorrespondingauthor{Nikita Doikov}{nikita.doikov@cornell.edu}
  \icmlcorrespondingauthor{Anastasia Koloskova}{anastasiia.koloskova@uzh.ch}

  \icmlkeywords{Certified Unlearning, Optimization Complexity, First-Order Methods, Second-Order Methods}

  \vskip 0.3in
]

\printAffiliationsAndNotice{}  %

\begin{abstract}
    We study machine unlearning: the removal of memorized training data from a trained model.
    Specifically, we investigate the algorithmic complexity of certified unlearning from an optimization perspective.
    We formalize the goal of an unlearning algorithm as simultaneously achieving certified unlearning and optimization accuracy.
    Utilizing the notion of uniformly convex regularizers,
    we prove new bounds on the distance between initial and unlearned models
    using a novel substitute 
    for generalization error. Thus we theoretically demonstrate that if the removed data is well-predicted by the unlearned model, the corresponding optimization problem is simple.
    Furthermore, we develop a new second-order unlearning algorithm
    with an anisotropic Gaussian mechanism
    and state-of-the-art global convergence.
    We prove fast rates for our method in achieving certified unlearning
    for linear models with quasi-self-concordant losses.
    As a direct application, our theory covers unlearning
    for logistic and exponential regressions and shows a provable benefit of utilizing second-order information
    compared to first-order unlearning methods.
\end{abstract}

\section{Introduction}

\subsection{Certified Unlearning and Optimization}

A crucial requirement for modern AI systems
is the ability to \textit{unlearn} the data,
that is to allow the data provider (e.g., an individual user or
an institutional organization) to select
\textit{which data is no longer allowed to be used}.
To satisfy ethical and legal conditions,
such prohibitions must be strictly enforced.
At the same time, retraining a large model from scratch
to forget a certain small portion of data 
is often very expensive or even impossible.
Therefore,
we are interested in efficient practical algorithms that 
enable removing the data from the model without retraining 
from scratch.
In this work,
we study the complexity of the machine unlearning
problem through the lens of optimization theory.
We develop a new \textit{second-order unlearning algorithm},
equipped with state-of-the-art fast global convergence rates
and a strong theoretically certified unlearning guarantee,
which ensures trustworthy models.

Let us denote by $A_N$ the entire dataset of size $N$ on which we have trained some machine learning model $\xx_F^\star$. The goal of unlearning is to forget a subset of the training data $A_N$ from our trained model $\xx_F^\star$. 
Denote by $A_n \subset A_N$ the subset of $A_N$ of size $n$ that remains after unlearning, frequently called the \textit{retain set}.
Therefore, our goal is to forget the data $A_N \setminus A_n$
of size $m = N - n$, and typically $m \lll N$
(removing a small portion of data).

Before specifying the certified unlearning precisely, first let us fix a \textit{learning algorithm}.
In this work, we analyze the unlearning problem from the optimization perspective.
Therefore, the process of learning consists of \textit{empirical risk minimization} (ERM).
We denote by $F_{\psi}(\cdot)$ the corresponding ERM objective on the whole dataset $A_N$ that we use to train an initial model, and by $f_{\psi}(\cdot)$
on the retain data $A_n$ (see definition of optimization formulations in Section~\ref{SectionOptimization}).

We assume that our initial model $\xx_F^\star$ is an exact solution to the ERM objective on the entire dataset $A_N$
\beq \label{OptSolutions1}
\ba{rcl}
\xx_F^{\star} & := & \argmin_{\xx} F_{\psi}(\xx). %
\ea
\eeq
This assumption follows the previous literature~\cite{sekhari2021remember}.
It provides a strong theoretical baseline, while it can be relaxed.
It will also be convenient to define the precise solution on the retain data $A_n$
\beq \label{OptSolutions2}
\ba{rcl}
\xx_f^{\star} & := & \argmin_{\xx} f_{\psi}(\xx).
\ea
\eeq
Knowing $\xx_f^{\star}$ provides an ideal solution to the unlearning problem, and we do not assume knowledge of $\xx_f^{\star}$.

We define an \textit{unlearning algorithm} as a randomized procedure 
$\yy = \U(\xx, A_N \setminus A_n, A_N)$ that takes as input, correspondingly,
a model $\xx$, the data $A_N \setminus A_n$ that we want to forget,
and the whole dataset $A_N$, or a set of possible statistics from it.
It returns a new unlearned model $\yy$. 

We use the following formal definition of \textit{certified unlearning}.
Let us fix a desired level $q \in (0, 1)$ and small $\delta > 0$.
We say that $\U$ is \textit{$(q, \delta)$-unlearning} if
for any measurable set~$Y$:
\beq \label{DefUnlearning}
\ba{rcl}
\P\bigl( \, \yy  \in Y \, \bigr)
& \leq & e^{q} \cdot \P\bigl(\, \yy^{\star} \in Y \, \bigr) + \delta \\[10pt]
\P\bigl( \, \yy^{\star}  \in Y \, \bigr)
& \leq & e^{q} \cdot \P\bigl(\, \yy \in Y \, \bigr) + \delta 
\ea
\eeq
where $\yy := \U(\xx_F^{\star}, A_N \setminus A_n, A_N)$ is a typical use of the unlearning algorithm, starting from the trained model $\xx_F^{\star}$ on the entire dataset,
and $\yy^{\star} := \U(\xx_f^{\star}, \varnothing, A_n)$ is the \textit{idealized theoretical use},
as if we train the model from scratch to obtain $\xx_f^{\star}$.
Note that this definition is aligned with the standard ones from the literature~\cite{koloskova2025certified, ginart2019making, guo2020certified}.
It is also closely related to the notion of \textit{differential privacy}~\cite{dwork2014algorithmic, feldman2022hiding}.

The naive mechanism to ensure~\eqref{DefUnlearning} consists of adding Gaussian noise 
to the output model of a training procedure~\cite{dwork2014algorithmic}.
In principle, \textit{any training algorithm} can be turned into $(q, \delta)$-unlearning,
if the variance of the added noise is sufficiently large.
However, such noise may erase not only the information to be forgotten, but also 
useful information about the retained data~$A_n$, yielding a model that is far from 
the desired solution~\eqref{OptSolutions2}. Hence, in this work,
along with the certified unlearning guarantee~\eqref{DefUnlearning}, we require
\textit{the output $\yy$ to be close to the exact minimum}:
\beq \label{OptGuarantee}
\ba{rcl}
\E\bigl[ \| \yy - \xx_f^{\star}  \| \bigr] & \leq & \epsilon,
\ea
\eeq
for a desired optimization tolerance $\epsilon > 0$.

\subsection{Contributions}

We establish rigorous complexity bounds
for optimization algorithms
to simultaneously achieve 
both
\textit{certified unlearning}~\eqref{DefUnlearning}
and \textit{optimization}~\eqref{OptGuarantee}
guarantees, for a given triplet $(q, \delta, \epsilon)$
of parameters specifying the problem.
To the best of our knowledge, we are the first
to establish fast global convergence
of second-order (Newton-type) methods for unlearning,
when the distance $\| \xx_F^{\star} - \xx_f^{\star}\|$ between models
can be arbitrarily far and propose a novel \textit{anisotropic Gaussian mechanism},
which is well suited to the geometry of the problem.
We summarize contributions as follows:

\begin{itemize}

\item Under assumption of uniform convexity of a model regularizer (see definition~\eqref{UConvex}; a particular case is $\ell_2$-regularization), we show how to relate the distance
between models $\xx_F^{\star}$ and $\xx_f^{\star}$, and the functional residual with 
a quantity $\pi(\xx)$ representing \textit{prediction error} of a model $\xx$ on unlearned data (Lemma~\ref{LemmaDistance}).
Thus, we show that if the unlearned model generalizes well on the removed data, then the distance is small,
and the corresponding optimization problem is simple.

\item We developed a \textit{new certified second-order unlearning method} (Algorithm~\ref{MainAlgorithm}).
It is based on computing the Hessian of the empirical loss, and using the state-of-the-art globalization of Newton's method for fast convergence even if $\xx_F^{\star}$ and $\xx_f^{\star}$ are far from each other. To ensure the certified unlearning,
we develop a novel mechanism of adding \textit{anisotropic} Normal distribution, which is naturally aligned with the Hessian.

\item Under the smoothness condition of \textit{quasi-self-concordance} of the loss~\cite{bach2010self,sun2019generalized,karimireddy2018global,doikov2025minimizing},
we show fast global convergence for our algorithm (Theorem~\ref{TheoremComplexityNewton}).
To the best of our knowledge, our complexity bound is also new to the optimization literature.
A working example that satisfies all our theoretical assumptions
is \textit{logistic} or \textit{exponential} regression with linear models, augmented by
\textit{any uniformly convex regularizer}. 
We prove the certified unlearning guarantee for our algorithm in Theorem~\ref{TheoremUnlearning}.

\end{itemize}

\subsection{Notation}

We fix some positive definite symmetric matrix $\BB \in \R^{d \times d}$, and define the primal-dual pair of generalized Euclidean norms, for any $\xx, \ss \in \R^d$:
\beq \label{GlobalNorm}
\ba{rcl}
\!\!\!\!
\| \xx \| & := & \la \BB \xx, \xx \ra^{1/2}, \;\;
\| \ss \|_* \;\; = \;\;
\la \ss, \BB^{-1}\ss \ra^{1/2}.
\ea
\eeq
We use the dual norm to measure the size of the gradients. In the simplest case, we can set $\BB := \II$ (identity matrix), which recovers the standard Euclidean norm. In general, matrix $\BB = \BB^{\top} \succ 0$ allows to better capture the geometry of the problem (see Proposition~\ref{PropositionApprox}). We also use $\BB$ for our novel sampling mechanism that ensures certified unlearning.

We say that a convex, not necessarily differentiable, regularizing function $\psi: \dom \psi \to \R$, 
is \textit{uniformly convex} of degree $p \geq 2$ with constant $\mu > 0$
(see, e.g., Chapter 4.2.2 in \cite{nesterov2018lectures})
if the symmetrized Bregman divergence is bounded as follows, for all $\xx, \yy \in \dom \psi$:
\beq \label{UConvex}
\ba{rcl}
\!\!\!\!\!\!
\bar{\beta}_\psi(\xx; \yy) & \!\!\!\! := \!\!\!\! 
& \la \psi'(\xx) - \psi'(\yy), \xx - \yy \ra 
\geq \mu \|\xx - \yy\|^p,
\ea
\eeq
where $\psi'(\xx) \in \partial \psi(\xx)$ is an arbitrary selection of subgradients.
Uniformly convex functions of degree $p = 2$ are called \textit{strongly convex}.

\section{Optimization Problem}
\label{SectionOptimization}

Let us consider the unlearning problem from the optimization perspective. 
We denote by $F$ the initial objective of \textit{training on the full dataset}:
\beq \label{InitProblem}
\ba{rcl}
\min\limits_{\xx \in \R^d} \Bigl[ \, F_{\psi}(\xx) 
\;\; = \;\; F(\xx) + \psi(\xx) \, \Bigr],
\ea
\eeq
augmenting it with a possible simple\footnote{Namely, we assume that we can efficiently solve a second-order subproblem involving~$\psi$ in our algorithm. The main example is $\ell_2$-regularization: $\psi(x) = \frac{\mu}{2}\|x\|^2$, which satisfies~\eqref{UConvex} with $p = 2$. Moreover, we can cover simple constraints in our model.} regularizer $\psi$.
We denote by $\xx_F^{\star} \in \R^d$ a solution to the initial training problem~\eqref{InitProblem}, which is available to us.
We use $\xx_F^{\star}$ as a \textit{starting point} for our unlearning algorithms. 
For simplicity, we assume that $\xx_F^{\star}$ is an exact minimizer to~\eqref{InitProblem},
while this assumption can be relaxed, using an approximate solution.

Further, we have the following decomposition of the initial objective, for some $0 \leq \gamma \leq 1$:
\beq \label{FBigDecomposition}
\ba{rcl}
F(\xx) & = & (1 - \gamma) f(\xx) + \gamma u(\xx),
\ea
\eeq
where $f: \R^d \to \R$ is the loss on \textit{core data}, that we keep in the dataset, and $u: \R^d \to \R$ is the loss on \textit{forget data}, that we are required to remove from training.
We assume that all training components, $F(\cdot)$, $f(\cdot)$, and $u(\cdot)$ are differentiable functions, while the regularizer $\psi$ can be non-differentiable (e.g. a mixture of $\ell_2$ and $\ell_1$-regularizers, or indicator of convex constraints).
Our main optimization objective consists of minimizing the following function on retain data:
\beq \label{MainProblem}
\ba{rcl}
\min\limits_{\xx \in \R^d} \Bigl[ \, f_{\psi}(\xx) 
\;\; = \;\; f(\xx) + \psi(\xx) \, \Bigr].
\ea
\eeq
We denote by $\xx_f^{\star} \in \R^d$ a solution to~\eqref{MainProblem},
which we want to find. 
To characterize the global complexity of solving~\eqref{MainProblem},
we introduce the quantity:
$
\pi(\xx) := \| \nabla u(\xx) + \psi'(\xx) \|_*,
$
which has an interpretation of a \textit{prediction error} of a model $\xx$ on the forget data $\nabla u$. 
In what follows, we show that this quantity plays the main role in the \textit{optimization complexity of algorithmic unlearning}. We can relate $\pi(\xx^{\star}_f)$ to the distance between solutions of two problems~\eqref{InitProblem} and~\eqref{MainProblem}, and $\pi(\xx^{\star}_F)$ to the functional residual:

\begin{lemma} \label{LemmaDistance}
	Let objectives $f$ and $F$ be convex with the relation defined in \eqref{FBigDecomposition}. Then,
	\beq \label{BoundDistance}
	\ba{rcl}
	\bar{\beta}_{\psi}(\xx^{\star}_{F}; \xx^{\star}_f) \cdot \| \xx_F^{\star} - \xx_f^{\star} \|^{-1}
	& \leq & 
	\gamma \pi(\xx^{\star}_f),
	\ea
	\eeq
	For uniformly convex regularizers~\eqref{UConvex}, we have
	\beq \label{BoundDistanceNorm}
	\ba{rcl}
	\| \xx^{\star}_F - \xx^{\star}_f \| & \leq &  
    \Bigl[ \frac{\gamma}{\mu} \pi(\xx^{\star}_f) \Bigr]^{\frac{1}{p - 1}}
	\ea
	\eeq
    and the bound for the functional residual:
    \beq \label{BoundFuncResidual}
    \ba{rcl}
    \!\!\!\!\!
    f_{\psi}(\xx^{\star}_F) - f_{\psi}(\xx^{\star}_f)
    & \!\!\! \leq \!\!\! &
    \frac{p - 1}{p}
    \Bigl[ 
    \frac{\gamma}{(1 - \gamma) \mu^{1/p}}  \pi(\xx^{\star}_F)
    \Bigr]^{\frac{p}{p - 1}}.
    \ea
    \eeq
\end{lemma}

Lemma~\ref{LemmaDistance} shows that if the unlearned model works well on the removed data (in other words, the removed data does not affect the generalization of the model and falls well within the distribution of the retain data), then the distance between the minimizers is also small.
We also conclude that the distance in the left hand side of~\eqref{BoundDistanceNorm} and the functional residual~\eqref{BoundFuncResidual} can be controlled by varying parameter $\gamma$ and~$\mu$. We can make the distance between the minimizers small both by setting $\gamma \to 0$ or by increasing the regularization parameter~$\mu > 0$.

\subsection{Examples and Assumptions}

Let $\aa_1, \ldots, \aa_n \in \R^d$ represent the \textit{core data}, and
$\aa_{n + 1}, \ldots, \aa_{n + m} \in \R^d$ be the \textit{forget data}.
We denote by $N = n + m$ the initial dataset size, 
and typically $n \ggg m$
(only a very small portion of the data is being removed).

\begin{example}[Linear Models] \label{ExampleLinear}
	Let $\ell : \R \to \R$ be a loss function. Consider
	the objective of training the generalized linear models:
	$
	F(\xx) :=
    \frac{1}{N} \sum_{i = 1}^N \ell( \la \aa_i, \xx \ra ),
	$
	and the corresponding \textit{core} and \textit{forget} components:
	\beq \label{LearnUnlearnComponents}
	\ba{rcl}
	f(\xx) & := & 
    \frac{1}{n} \sum\limits_{i = 1}^n \ell( \la \aa_i, \xx \ra ), \\
	u(\xx) & := &
	\frac{1}{m} \sum\limits_{j = n + 1}^{n + m} \ell( \la \aa_j, \xx \ra ).
	\ea
	\eeq
	Then, decomposition~\eqref{FBigDecomposition} holds with $\boxed{\gamma = \tfrac{m}{N}}$. 
    Note that $\gamma \to 0$ when $N \ggg m$,
    which is a common scenario.
\end{example}

We assume that the loss function $\ell: \R \to \R$ is convex, differentiable, and sufficiently smooth. In particular, we use the following notion of smoothness which captures the local geometry of the objective~\cite{bach2010self}:

\begin{assumption}[Quasi-Self-Concordance] 
	Assume that it holds, for some constant $M \geq 0$:
	\beq \label{QSC}
	\ba{rcl}
	|\ell'''(t)| & \leq & M \cdot \ell''(t), \qquad t \in \R.
	\ea
	\eeq
\end{assumption}

\begin{example}[Quasi-Self-Concordant Losses] \label{ExampleLosses}
The following functions satisfy assumption~\eqref{QSC}:
\begin{itemize}
	\item \textit{Quadratic loss}, $\ell(t) = \frac{1}{2}t^2$. Then $M = 0$.
	\item \textit{Exponential loss}, $\ell(t) = e^t$. Then $M = 1$.
	\item \textit{Logistic loss}, for classification of two (or more) classes, which can be written as, $\ell(t) = \log(1 + e^t)$. Then $M = 1$.
\end{itemize}
\end{example}

The parameter $M \geq 0$ measures how far the loss is from a quadratic function. Then, to capture the geometry of the retained model~\eqref{LearnUnlearnComponents}, we define
\beq \label{BChoice}
\boxed{
\ba{rcl}
\BB & := & \sum\limits_{i = 1}^n \aa_i \aa_i^{\top},
\ea
}
\eeq
which defines the global norm~\eqref{GlobalNorm}
and anisotropic Gaussian mechanism
that we use in the method.
This choice ensures data-agnostic second-order approximation of the objective~\cite{doikov2025minimizing}. 
For non-linear models, a suitable choice is the Hessian at the initialization: $\BB := \nabla^2 f(\xx_0)$,
which approximates~\eqref{BChoice}.

\section{Algorithm}

At each iteration $k \geq 0$ of our algorithm, we use a positive definite matrix $\HH_k = \HH_k^{\top} \succ 0$
that is designed to capture a second-order information about objective and accelerate the global unlearning.
Using this matrix, we define the second-order model $m_k$ around $\xx_k$, with regularizer:
$$
\ba{rcl}
\!\!\!
m_k(\xx) & \!\!\!\!\! := \!\!\!\!\! & \la \nabla f(\xx_k), \xx \ra 
\! + \! \frac{1}{2}\la \HH_k(\xx - \xx_k), \xx - \xx_k \ra \! + \! \psi(\xx).
\ea
$$
We discuss how to choose the matrix $\HH_k$ in the next sections.
In our algorithm, we minimize this model for $K \geq 1$ iterations, starting from
the minimum $\xx^{\star}_F$ of the full model~\eqref{InitProblem}.
After that, to reach a certified unlearning guarantee,
we apply anisotropic Gaussian mechanism to the output.

\renewcommand{\algorithmicrequire}{\textbf{Init:}}
\renewcommand{\algorithmicfor}{\textbf{For}}
\renewcommand{\algorithmicendfor}{\textbf{End for}}

\begin{algorithm}[ht]
\caption{Certified Second-Order Unlearning}
\label{MainAlgorithm}
\begin{algorithmic}[1]
\REQUIRE Set $\xx_0 := \xx_F^{\star}$. 
Fix $\BB \succ 0$, $\sigma > 0$, and $K \geq 1$.
\FOR{$k = 0 \ldots K - 1$}
\STATE Compute next step $\xx_{k + 1} = \argmin_{\xx}\bigl[ m_k(\xx) \bigr]$ 
\ENDFOR
\STATE Sample a normal vector $\xixi \sim \mathcal{N}(\vzero, \sigma^2 \BB^{-1}) $
\STATE \textbf{Return} $\yy_K = \xx_K + \xixi$
\end{algorithmic}
\end{algorithm}

For convenience, we denote the norm of the current (sub)gradient at each iterate by
$g_k := \| \nabla f(\xx_k) + \psi'(\xx_k) \|_*$.
We analyze two instances of our algorithm.

\subsection{Gradient Method Baseline}

This is the first-order baseline of our approach, which selects
$\BB := \II$ (isotropic Gaussian noise) and $\HH_k := \eta \II$, for a certain step-size parameter $\eta > 0$.
In case $\psi(x) \equiv 0$ (no regularization), iterations of our algorithm read as
$$
\ba{rcl}
\xx_{k + 1} & = &
\xx_k - \frac{1}{\eta} \nabla f(\xx_k),
\ea
$$
which is the standard gradient descent.
For general $\psi$, each iteration can be represented through the prox operator of $\psi$.

\begin{theorem} \label{TheoremComplexityGM}
Assume that the second derivative of the loss is bounded: $L \geq \ell''(t), \forall t$.
We set $\eta := L\|\AA \|^2$, where $\AA \in \R^{n \times d}$ 
is the matrix of the retained data.
Then, we achieve $\|\xx_K - \xx^{\star}_f\| \leq \varepsilon$
after the following number of the gradient steps, for $p > 2$:
    \beq \label{GradMethodComplexity}
	\ba{rcl}
    \!\!\!\!
	K & \!\! = \!\! & 
    O\Bigl(\, 
    \frac{L \| \AA \|^2}{\mu^{2 / p}}
    \Bigl[\, \bigl(\frac{1}{\mu \varepsilon^p}\bigr)^{\frac{p - 2}{p}}  
    - \bigl( \frac{\mu^{1/p}}{\gamma \pi(\xx^{\star}_F)} \bigr)^{\frac{p - 2}{p - 1}} \,\Bigr]
    \,\Bigr),
	\ea
	\eeq
    and, for $p = 2$ (strongly convex case):
    \beq \label{GradMethodStronglyConvex}
    \ba{rcl}
    K & = & O\Bigl(\, \frac{L\| \AA \|^2}{\mu} \ln\frac{ \gamma \pi(\xx^{\star}_F) }{\mu \varepsilon} \,\Bigr).
    \ea
    \eeq
\end{theorem}

\subsection{Newton Method with Gradient Regularization}

In our most advanced second-order instance of the algorithm, we choose,
as in~\cite{doikov2024super,doikov2025minimizing}:
\beq \label{MatrixChoice}
\boxed{
\ba{rcl}
\HH_k & := & \nabla^2 f(\xx_k) + M g_k \BB
\ea
}
\eeq
where $M$ is a quasi-self-concordant parameter of the loss, 
and $\BB = \sum_{i = 1}^n \aa_i \aa_i^{\top}$
(see Proposition~\ref{PropositionApprox}).
The most important case  is when $\psi(\yy) = \frac{\mu}{2}\| \yy \|^2$ (strongly-convex regularizer, $p = 2$). Then, each iteration of our method can be written explicitly, as follows, for $k \geq 0$:
$$
\ba{rcl}	
\!\!
\xx_{k + 1} &\!\!\! = \!\!\!\! & \xx_k - 
\Bigl( \nabla^2 f(\xx_k) + (M g_k + \mu) \BB  \Bigr)^{-1} \nabla f_{\psi}(\xx_k),
\ea
$$
and for $M = 0$ this is the classical Newton method as applied to~\eqref{MainProblem}.
By employing the gradient regularization,
we ensure \textit{fast global convergence} of our algorithm, as shown in the following theorem.
Note that for a general regularizer $\psi$, the model $m_k(\yy)$ is strongly convex due to~\eqref{MatrixChoice},
and can be solved efficiently by first-order optimization subroutines,
without additional data samples.

\begin{theorem} \label{TheoremComplexityNewton}
    For quasi-self-concordant loss and uniformly convex regularizer, we achieve
	$\| \xx_K - \xx^{\star}_f \|  \leq  \varepsilon$,
	after the following number of the Newton steps with~\eqref{MatrixChoice}:
	\beq \label{NewtonComplexity}
	\ba{rcl}
	K & = & O\Bigl( M \Bigl[ \frac{\gamma}{\mu} \pi(\xx^{\star}_F) \Bigr]^{1/p} 
    + \log\frac{\gamma \pi(\xx_F^{\star})}{\mu \varepsilon^{p - 1}} \Bigr).
	\ea
	\eeq
\end{theorem}

We see that, in contrast to the gradient method,
the complexity~\eqref{NewtonComplexity} of the Newton method
is much better, as $\varepsilon > 0$ enters as an additive logarithmic term.
Moreover, the rate of the Newton method does 
not depend on the size of the input data $\| \AA \|^2$ as 
in~\eqref{GradMethodComplexity} and~\eqref{GradMethodStronglyConvex}.
The main complexity factor is the first term, which is small
when either $\gamma \to 0$ (unlearning a small portion of data)
or $\pi(\xx^{\star}_F)$ is small (good prediction of the full model on unlearned data).

\subsection{Certified Unlearning Guarantee}

Applying anisotropic Gaussian mechanism in the output of Algorithm~\ref{MainAlgorithm}, 
we are able to prove our main result on certified unlearning.

\begin{theorem} \label{TheoremUnlearning}
Let $\BB \succ 0$ be arbitrary and 
assume that $\xx_K$ satisfies 
the optimization guarantee 
$\| \xx_K - \xx^{\star}_f \| \leq \varepsilon$.
For any $q \in (0, 1)$ and $\delta > 0$,
set 
\beq \label{SigmaVal}
\ba{rcl}
\sigma & := & \frac{\varepsilon}{q} \max\Bigl\{ 1, 2 \sqrt{2 \ln \frac{2}{\delta}} \Bigr\}.
\ea
\eeq
Then, Algorithm~\ref{MainAlgorithm} ensures certified $(q, \delta)$-unlearning.
Moreover, the result satisfies the optimization guarantee:
\beq \label{BoundSolution}
\ba{rcl}
\E\Bigl[ \| \yy_K - \xx^{\star}_f \|  \Bigr]
& \leq & \sqrt{\varepsilon^2 + \sigma^2 d}.
\ea
\eeq
\end{theorem}

\begin{corollary}
    For a given triplet of parameters $(q, \delta, \epsilon)$,
    Algorithm~\ref{MainAlgorithm} achieves $(q, \delta)$-unlearning~\eqref{DefUnlearning} and $\epsilon$-bound for optimization guarantee~\eqref{OptGuarantee} in total of
    $$
    \ba{rcl}
    K & \!\!\!\! = \!\!\!\! & 
    O\Bigl( M\Bigl[ \frac{\gamma}{\mu} \pi(\xx^{\star}_F)  \Bigr]^{1 / p} 
    + \log\Bigl[ \frac{\gamma \pi(\xx^{\star}_F) }{\mu}
      \Bigl( \frac{d \ln 1/\delta}{q \epsilon^2} \Bigr)^{\frac{p - 1}{2}}
         \Bigr] \Bigr)
    \ea
    $$
    iterations (retained data passes). Therefore, we see that all
    key parameters $(q, \delta, \epsilon)$ enter under the logarithm,
    and the efficiency depends mainly
    on $\gamma$ (the portion of removed data), $\mu > 0$ (regularization coefficient), and $\pi(\xx^{\star}_F)$.
\end{corollary}

\newpage


\newpage
\appendix
\onecolumn

\section{Proofs}

The main consequence of condition~\eqref{QSC} that we use is the following bound for the Hessian of the linear models
(see Lemma~2 in~\cite{doikov2025minimizing} for the proof):

\begin{proposition} \label{PropositionApprox}
	Consider generalized linear models~\eqref{LearnUnlearnComponents} with quasi-self-concordant loss~\eqref{QSC}. Then,
	\beq \label{QSCFunc}
	\ba{rcl}
	\| \nabla f(\yy) - \nabla f(\xx) - \nabla^2 f(\xx)(\yy - \xx) \|_*
	& \leq  & M \|\yy - \xx\|_{\xx}^2 \cdot \varphi( M \|\yy - \xx\| ),
	\ea
	\eeq
	where $\varphi(t) := \frac{e^t - 1 - t}{t^2} > 0$ is a convex monotone function, $\|\yy - \xx\|_{\xx} 
    = \la \nabla^2 f(\xx)(\yy - \xx), \yy - \xx \ra^{1/2}$ is the local norm induced by the Hessian of the unlearning problem,
    and $$
    \ba{rcl}
    \BB & = & \sum\limits_{i = 1}^n \aa_i \aa_i^{\top},
    \ea
    $$
    as in~\eqref{BChoice},
    defines the global norm~\eqref{GlobalNorm}
    that we use in the method.
\end{proposition}

\subsection{Proof of Lemma~\ref{LemmaDistance}}

The optimality condition for the 
minimizer~$\xx^{\star}_F$  of the initial model~\eqref{InitProblem} is
\beq \label{OptConditionF}
\ba{rcl}
\nabla F(\xx^{\star}_F) + \psi'(\xx^{\star}_F) & = & 0 
\qquad \text{where} \qquad \psi'(\xx^{\star}_F) \;\; \in \;\; \partial \psi(\xx^{\star}_F).
\ea
\eeq
At the same time, the optimality condition for the minimizer $\xx^{\star}_f$ 
of the unlearned model~\eqref{MainProblem} is
\beq \label{OptConditionf}
\ba{rcl}
\nabla f(\xx^{\star}_f) + \psi'(\xx^{\star}_f) & = & 0
\qquad \text{where} \qquad  
\psi'(\xx^{\star}_f) \;\; \in \;\; \partial \psi(\xx^{\star}_f).
\ea
\eeq
By convexity of $F$, we have
$$
\ba{rcl}
\bar{\beta}_{\psi}(\xx^{\star}_F; \xx^{\star}_f)
& = & 
\la \psi'(\xx^{\star}_F) - \psi'(\xx^{\star}_f), \xx^{\star}_F - \xx^{\star}_f \ra \\
\\
& \leq &
\la \nabla F(\xx^{\star}_F) + \psi'(\xx^{\star}_F) - \nabla F(\xx^{\star}_f) - \psi'(\xx^{\star}_f), \xx^{\star}_F - \xx^{\star}_f \ra \\
\\
& \overset{(\ref{OptConditionF})}{=} &
\la \nabla F(\xx^{\star}_f) + \psi'(\xx^{\star}_f), \xx^{\star}_f - \xx^{\star}_F \ra \\
\\
& \overset{(\ref{FBigDecomposition})}{=} &
(1 - \gamma) \la \nabla f(\xx^{\star}_f) + \psi'(\xx^{\star}_f), \xx^{\star}_f - \xx^{\star}_F \ra
+ \gamma \la \nabla u(\xx^{\star}_f) + \psi'(\xx^{\star}_f), \xx^{\star}_f - \xx^{\star}_F \ra \\
\\
& \overset{(\ref{OptConditionf})}{=} &
\gamma \la \nabla u(\xx^{\star}_f) + \psi'(\xx^{\star}_f), \xx^{\star}_f - \xx^{\star}_F \ra
\;\; \leq \;\;
\gamma \| \nabla u(\xx^{\star}_f) + \psi'(\xx^{\star}_f) \|_* \cdot \| \xx^{\star}_f - \xx^{\star}_F \|.
\ea
$$
Rearranging the terms gives~\eqref{BoundDistance}. 

Then, applying the uniform convexity~\eqref{UConvex} to~\eqref{BoundDistance} yields~\eqref{BoundDistanceNorm}. 

Finally, since $f_{\psi}(\xx) = f(\xx) + \psi(\xx)$ is uniformly convex as a sum of a convex function $f$ and uniformly convex regularizer~$\psi$, by simple integration we obtain, for any $\xx, \yy \in \dom \psi$ and $f'_{\psi}(\xx) \in \partial f_{\psi}(\xx)$:
\beq \label{UConvFunc}
\ba{rcl}
f_{\psi}(\yy) & \geq & f_{\psi}(\xx) + \la f'_{\psi}(\xx), \yy - \xx \ra + \frac{\mu}{p}\|\yy - \xx\|^p.
\ea
\eeq
Minimizing the left and the right hand side independently with respect to $y$ gives:
\beq \label{FuncResUConvex}
\ba{rcl}
f_{\psi}(\xx) - f_{\psi}(\xx^{\star}_f) & \leq & 
\frac{p - 1}{p} \frac{ \| f'_{\psi}(\xx) \|^{\frac{p}{p - 1}}}{\mu^{\frac{1}{p - 1}}},
\qquad \xx \in \dom \psi.
\ea
\eeq
It remains to substitute $\xx := \xx^{\star}_F$ and notice that, due to~\eqref{FBigDecomposition},
$\| f'_{\psi}(\xx^{\star}_F) \|_* = \frac{\gamma}{1 - \gamma} \|\nabla u(\xx^{\star}_F) + \psi'(\xx^{\star}_F)  \|_*
= \frac{\gamma}{1 - \gamma}\pi(\xx^{\star}_F)$, which completes the proof.
\qed

\subsection{Proof of Theorem~\ref{TheoremComplexityNewton}}

Let us denote by $f_k := f_{\psi}(\xx_k) - f_{\psi}(\xx^{\star}_f) \geq 0$
the functional residual at iteration $k \geq 0$.
Then, for one step of the Newton method with gradient regularization, we have the following progress (see Theorem~3.2 in~\cite{doikov2025minimizing} for $(*)$), employing additionally uniform convexity of the regularizer:
\beq \label{Progress}
\ba{rcl}
f_k - f_{k + 1} & \overset{(*)}{\geq} & 
\frac{1}{2M} \Bigl[ \frac{g_{k + 1}}{g_k} \Bigr]^2 g_k
\;\; \overset{(\ref{FuncResUConvex})}{\geq} \;\;
c \cdot \Bigl[ \frac{g_{k + 1}}{g_k} \Bigr]^2 f_k^{\frac{p - 1}{p}},
\ea
\eeq
where $c := \frac{\mu^{1 / p}}{2M} \Bigl[ \frac{p}{p - 1} \Bigr]^{\frac{p - 1}{p}}$.

Note that due to concavity of $\varphi(t) = t^{1 / p}$, we have, for any $a, b > 0$:
\beq \label{Concave1p}
\ba{rcl}
b^{1/p} & \leq & a^{1/p} + \frac{1}{p} a^{-\frac{p - 1}{p}} (b - a)
\qquad \Leftrightarrow \qquad
a^{1/p} - b^{1/p} \;\; \geq \;\;
\frac{1}{p} a^{-\frac{p - 1}{p}} (a - b).
\ea
\eeq
Therefore, we have
\beq \label{ProgressNormalized}
\ba{rcl}
f_k^{1 / p} - f_{k + 1}^{1 / p}
& \overset{(\ref{Concave1p})}{\geq} &
\frac{1}{p} f_k^{- \frac{p - 1}{p}} ( f_k - f_{k + 1} )
\;\; \overset{(\ref{Progress})}{\geq} \;\;
\frac{c}{p} \Bigl[ \frac{g_{k + 1}}{g_k} \Bigr]^{2}.
\ea
\eeq
Telescoping this progress for the first $K \geq 1$ iterations,
and using the inequality between arithmetic and geometric means, we get
\beq \label{TelescopedProgress}
\ba{rcl}
f_0^{1/p} - f_K^{1/p}
& \overset{(\ref{ProgressNormalized})}{\geq} &
\frac{cK}{p} \cdot \frac{1}{K} \sum\limits_{i = 0}^{K - 1} \Bigl[ \frac{g_{i + 1}}{g_i} \Bigr]^2
\;\; \geq \;\;
\frac{cK}{p} \cdot 
\Bigl[ \, \prod\limits_{i = 0}^{K - 1} \frac{g_{i + 1}}{g_i} \, \Bigr]^{2 / K}
\;\; = \;\;
\frac{cK}{p} \cdot \Bigl[ \, \frac{g_K}{g_0}  \, \Bigr]^{2 / K} \\
\\
& = &
\frac{cK}{p} \cdot \exp\Bigl( \frac{2}{K} \log \frac{g_K}{g_0} \Bigr)
\;\; \geq \;\;
\frac{cK}{p} \cdot \Bigl(1 + \frac{2}{K} \log \frac{g_K}{g_0}\Bigr).
\ea
\eeq
Using that $f_K \geq 0$ and rearranging the terms, we obtain
\beq \label{KBoundNewton}
\ba{rcl}
K & \overset{(\ref{TelescopedProgress})}{\leq} &
\frac{p}{c} f_0^{1/p}
+ 2 \log \frac{g_0}{g_K}
\;\; = \;\;
\frac{2p M}{\mu^{1/p}}\Bigl[ \frac{p - 1}{p} \Bigr]^{\frac{p - 1}{p}}
f_0^{1/p}
+ 2 \log \frac{g_0}{g_K}.
\ea
\eeq
Assuming that $\|\xx_K - \xx^{\star}_f\| \geq \varepsilon$ and using uniform convexity, we obtain the lower bound on the current (sub)gradient norm:
$$
\ba{rcl}
g_K & = & \| f'_{\psi}(\xx_K) \|_*
\;\; \overset{(\ref{FuncResUConvex}),(\ref{UConvFunc})}{\geq} \;\;
\frac{\mu}{(p - 1)^{\frac{p - 1}{p}}} \|\xx_K - \xx^{\star}_f \|^{p - 1}
\;\; \geq \;\;
\frac{\mu}{(p - 1)^{\frac{p - 1}{p}}} \varepsilon^{p - 1}.
\ea
$$
Substituting this bound into~\eqref{KBoundNewton}, and using 
the bound on the initial functional residual:
\beq \label{Func0Upper}
\ba{rcl}
f_0 & = & f_{\psi}(\xx^{\star}_F) - f_{\psi}(\xx^{\star}_f)
\;\; \overset{(\ref{BoundFuncResidual})}{\leq} \;\;
\frac{p - 1}{p}
\Bigl[ \frac{\gamma}{(1 - \gamma) \mu^{1/p}} \pi(\xx^{\star}_F) \Bigr]^{\frac{p}{p - 1}}
\ea
\eeq
completes the proof. \qed

\subsection{Proof of Theorem~\ref{TheoremComplexityGM}}

Since for analyzing the gradient method, we assume that the loss function
has bounded second derivative: $\ell''(t) \leq L$, $\forall t$, we conclude 
that the smooth part $f$ of the objective in~\eqref{MainProblem}
has Lipschitz continuous gradient with constant $L_f := L \| \AA \|^2$,
where $\AA \in \R^{n \times d}$ 
is the matrix composed by the retaining data $\aa_1, \ldots, \aa_n \in \R^d$.
In the gradient method, we use the classical choice of the stepsize parameter (see, e.g.~\cite{nesterov2018lectures}), as
\beq \label{EtaChoice}
\boxed{
\ba{rcl}
\eta & := & L_f \;\; = \;\; L \| \AA \|^2.
\ea
}
\eeq
Optimality condition for one method step, taking into account the regularizer~$\psi$, is
\beq \label{OptCondition}
\ba{rcl}
\nabla f(\xx_k) + \eta(\xx_{k + 1} - \xx_k)
+ \psi'(\xx_{k + 1}) & = & 0,
\qquad \psi'(\xx_{k + 1}) \;\; \in \;\; \partial \psi(\xx_{k + 1}).
\ea
\eeq
Therefore, using Lipschitzness of the gradient, we conclude that
$$
\ba{rcl}
L_f\|\xx_{k + 1} - \xx_k\|
& \geq & 
\| \nabla f(\xx_{k + 1}) - \nabla f(\xx_k) \|
\;\; \overset{(\ref{OptCondition}), (\ref{EtaChoice})}{=} \;\;
\| f'_{\psi}(\xx_{k + 1}) + L_f (\xx_{k + 1} - \xx_k) \|.
\ea
$$
Taking the square of both sides and rearranging the terms, we obtain
\beq \label{GMGradProgress}
\ba{rcl}
\la f'_{\psi}(\xx_{k + 1}), \xx_k - \xx_{k + 1}
\ra & \geq & \frac{1}{2L_f} \| f'_{\psi}(\xx_{k + 1}) \|^2
\;\; = \;\; \frac{1}{2L_f} g_{k + 1}^2.
\ea
\eeq
Then, for the functional residual $f_k := f_{\psi}(\xx_k) - f_{\psi}(\xx_f^{\star}) \geq 0$
and employing the uniform convexity of the regularizer, we obtain the recurrence:
\beq \label{GMProgress}
\ba{rcl}
f_k - f_{k + 1} & \geq &
\la f'_{\psi}(\xx_{k + 1}), \xx_k - \xx_{k + 1} \ra
\;\; \overset{(\ref{GMGradProgress})}{\geq} \;\;
\frac{1}{2L_f} g_{k + 1}^2
\;\; \overset{(\ref{FuncResUConvex})}{\geq} \;\;
c \cdot f_{k + 1}^{\frac{2(p - 1)}{p}}
\;\; = \;\;
c \cdot f_{k + 1}^{\alpha},
\ea
\eeq
where $c := \bigl( \frac{p}{p - 1} \bigr)^{\frac{2(p - 1)}{p}} \frac{\mu^{2 / p}}{2 L_f}$
and $\alpha := \frac{2(p - 1)}{p} \in [1, 2)$.

Note that the function $\varphi(t) = t^{\alpha - 1}$ is concave. Thus, for any $a, b > 0$:
\beq \label{PhiAlphaBound}
\ba{rcl}
b^{\alpha - 1} & \leq & a^{\alpha - 1}
+ (\alpha - 1) a^{\alpha - 2}(b - a)
\qquad \Leftrightarrow \qquad
\frac{1}{\alpha - 1}\Bigl[ a^{\alpha - 1} - b^{\alpha - 1} ] \;\; \geq \;\; a^{\alpha - 2}(a - b),
\ea
\eeq
where we treat the left hand side of the last expression as the limit when $\alpha \to 1$:
$$
\ba{rcl}
\lim\limits_{\alpha \to 1}
\frac{1}{\alpha - 1}\Bigl[ a^{\alpha - 1} - b^{\alpha - 1} ]  
& = & \log\frac{a}{b}.
\ea
$$
Hence, we obtain
\beq \label{OneStepGMProgress}
\ba{rcl}
\frac{1}{\alpha - 1}\Bigl[ \frac{1}{f_{k + 1}^{\alpha - 1}} - \frac{1}{f_k^{\alpha - 1}} \Bigr]
& = &
\frac{1}{f_{k + 1}^{\alpha - 1} f_k^{\alpha - 1} (\alpha - 1) }
\Bigl[ f_k^{\alpha - 1} - f_{k + 1}^{\alpha - 1} \Bigr]
\;\; \overset{(\ref{PhiAlphaBound})}{\geq} \;\;
\frac{f_k - f_{k + 1}}{f_{k + 1}^{\alpha - 1} f_k}
\;\; \overset{(\ref{GMProgress})}{\geq} \;\;
c \cdot \frac{f_{k + 1}}{f_k}.
\ea
\eeq
Telescoping this inequality for the first $K \geq 1$ iterations, and using the inequality between
arithmetic and geometric means, we get
$$
\ba{rcl}
\frac{1}{\alpha - 1}\Bigl[ \frac{1}{f_{K}^{\alpha - 1}} - \frac{1}{f_0^{\alpha - 1}} \Bigr]
& \overset{(\ref{OneStepGMProgress})}{\geq} &
cK \cdot \frac{1}{K} \sum\limits_{i = 0}^{K - 1} \frac{f_{i + 1}}{f_i}
\;\; \geq \;\;
cK \cdot \Bigl[\; \prod\limits_{i = 0}^{K - 1} \frac{f_{i + 1}}{f_i} \;\Bigr]^{1 / K}
\;\; = \;\;
cK \cdot \Bigl[\; \frac{f_K}{f_0} \;\Bigr]^{1 / K} \\
\\
& = &
cK \cdot \exp\Bigl(\; \frac{1}{K} \log \frac{f_K}{f_0}  \;\Bigr)
\;\; \geq \;\;
cK \cdot \Bigl(\; 1 + \frac{1}{K} \log \frac{f_K}{f_0}  \; ).
\ea
$$
Rearranging the terms, we have
\beq \label{KGMBound}
\ba{rcl}
K & \leq & 
\frac{1}{c(\alpha - 1)} \Bigl[ \frac{1}{f_{K}^{\alpha - 1}} - \frac{1}{f_0^{\alpha - 1}} \Bigr]
+ \log \frac{f_0}{f_K} \\
\\
& = &
2p \cdot \Bigl( \frac{p - 1}{p} \Bigr)^{\frac{2(p - 1)}{p}}
\cdot 
\frac{L \| \AA \|^2}{\mu^{2/p}}
\cdot \frac{1}{p - 2}\Bigl[\; 
\bigl(\frac{1}{f_K}\bigr)^{\frac{p - 2}{p}} - 
\bigl(\frac{1}{f_0}\bigr)^{\frac{p - 2}{p}}  \;\Bigr]
+ \log \frac{f_0}{f_K}.
\ea
\eeq
It remains to use the upper bound~\eqref{Func0Upper} on the initial functional residual $f_0$, and the lower bound on $f_K$,
assuming that $\|\xx_K - \xx_f^{\star} \| \geq \varepsilon$:
$$
\ba{rcl}
f_K & = & f_{\psi}(\xx_K) - f_{\psi}(\xx_f^{\star})
\;\; \overset{(\ref{UConvFunc})}{\geq} \;\;
\frac{\mu}{p} \| \xx_K - \xx_f^{\star} \|^p
\;\; \geq \;\;
\frac{\mu}{p} \varepsilon^p.
\ea
$$
Substituting these estimates into~\eqref{KGMBound} completes the proof.
\qed

\subsection{Proof of Theorem~\ref{TheoremUnlearning}}

Our proof follows the standard reasoning used in Gaussian mechanism for
differential privacy and unlearning (see, e.g., \cite{nikolov2013geometry, dwork2014algorithmic}).
Since we analyze it from the optimization perspective, and, additionally, in Algorithm~\ref{MainAlgorithm} we employ \textit{anisotropic Normal distribution},
which is better suitable for the second-order geometry, we provide the full proof for completeness of our presentation.

Let $\yy_K = \xx_K + \xixi$ be the output of Algorithm~\ref{MainAlgorithm}
starting from $\xx_0 := \xx_F^{\star}$,
where $\xixi \sim \mathcal{N}(\vzero, \sigma^2 \BB^{-1})$ is the Normal noise.

At the same time, note that if we run Algorithm~\ref{MainAlgorithm} from $\xx_0 := \xx_f^{\star}$,
which satisfies the optimality condition:
\beq \label{XfStarCondition}
\ba{rcl}
\nabla f(\xx_f^{\star}) + \psi'(\xx_f^{\star}) & = & 0, \qquad \psi'(\xx_f^{\star}) \in \partial \psi(\xx_f^{\star}),
\ea
\eeq
then all iterates are the same: $\xx_{K} = \xx_{K - 1} = \ldots = \xx_0 = \xx_f^{\star}$ (so $\xx_f^{\star}$
is a fixed point of the iterates).
Indeed, due to $\HH_k \succ 0$, every next iterate $\xx_{k + 1}$ is a unique solution of the following equation:
\beq \label{NextIterate}
\ba{rcl}
\nabla f(\xx_k) + \HH_k(\xx_{k + 1} - \xx_k) + \psi'(\xx_{k + 1}) & = & 0,
\qquad \psi'(\xx_{k + 1}) \; \in \; \partial \psi(\xx_{k + 1}),
\ea
\eeq
and it is easy to see that if $\xx_k = \xx_f^{\star}$, which satisfies~\eqref{XfStarCondition},
then $\xx_{k + 1} := \xx_f^{\star}$ satisfies~\eqref{NextIterate}.
Hence, running Algorithm~\ref{MainAlgorithm} from $\xx_0 := \xx_f^{\star}$
for any number of iterations $K \geq 0$, we always have as the result $\xx_K = \xx_f^{\star}$.
Let us denote the output of Algorithm~\ref{MainAlgorithm} in this case by 
$\yy^{\star} := \xx_f^{\star} + \xixi$, where $\xixi \sim \mathcal{N}(\vzero, \sigma^2 \BB^{-1})$.

Therefore, to show certified ($q, \delta$)-unlearning, by definition,
we need to show for any measurable $Y \subseteq \R^d$ that
\beq \label{UnlearningBounds}
\ba{rcl}
\P\bigl( \yy_K \in Y \bigr)
& \leq & 
e^{q} \cdot \P\bigl( \yy^{\star} \in Y \bigr) + \delta, \\
\\
\P\bigl( \yy^{\star} \in Y \bigr)
& \leq & 
e^{q} \cdot \P\bigl( \yy_K \in Y \bigr) + \delta.
\ea
\eeq
To establish~\eqref{UnlearningBounds}, we follow the reasoning from~\cite{nikolov2013geometry},
extending it to our case.
We denote $\vv := \xx_K - \xx_f^{\star}$, and by our optimization guarantee, we have 
\beq \label{VBounded}
\ba{rcl}
\|\vv \| & := & \la \BB \vv, \vv \ra^{1/2} \;\; \leq \;\; \varepsilon.
\ea
\eeq

Let $p(\xixi)$ be the probability density function of $\mathcal{N}(\vzero, \sigma^2 \BB^{-1})$:
$$
\ba{rcl}
p(\xixi) & \propto & \exp\Bigl( - \frac{1}{2\sigma^2} \la \BB \xixi, \xixi \ra \Bigr)
\;\; = \;\;
\exp\Bigl( - \frac{1}{2\sigma^2}\| \xixi \|^2 \Bigr),
\ea
$$
and consider the quantity:
\beq \label{DUpperBound}
\ba{rcl}
D_{\vv}(\xixi)
& := &
\ln \frac{p(\xixi)}{p(\xixi + \vv)}
\;\; = \;\;
\ln p(\xixi) - \ln p(\xixi + \vv)
\;\; = \;\;
\frac{1}{2\sigma^2} \| \xixi + \vv \|^2 - \frac{1}{2\sigma^2} \| \xixi \|^2 \\
\\
& = & 
\frac{1}{2\sigma^2} \| \vv \|^2 + \frac{1}{\sigma^2} \la \BB\vv, \xixi \ra
\;\; \overset{(\ref{VBounded})}{\leq} \;\;
\frac{\varepsilon^2}{2\sigma^2}  + \frac{1}{\sigma^2} \la \BB\vv, \xixi \ra.
\ea
\eeq
Note that 
$t := \frac{1}{\sigma^2} \la \BB\vv, \xixi \ra \sim \mathcal{N}\bigl(0, \frac{\| \vv \|^2}{\sigma^2}\bigr)$
is a univariate Normal variable. The classic Chernoff bound ensures that
\beq \label{TBound}
\ba{rcl}
\P\bigl( |t| \, \geq \, \alpha \bigr)
\;\; = \;\;
\P\bigl( \frac{1}{\sigma^2} |\la \BB\vv, \xixi \ra|  \, \geq \, \alpha  \bigr)
& \leq & 
2\exp\Bigl( - \frac{\alpha^2 \sigma^2}{2\| \vv \|^2} \Bigr)
\;\; \overset{(\ref{VBounded})}{\leq} \;\;
\delta
\;\; := \;\; 
2\exp\Bigl( - \frac{\alpha^2 \sigma^2}{2\varepsilon^2} \Bigr).
\ea
\eeq
Hence, with probability that is greater than $1 - \delta$, we have
\beq \label{DBoundQ}
\ba{rcl}
|D_{\vv}(\xixi)|
& \overset{(\ref{DUpperBound})}{\leq} &
\frac{\varepsilon^2}{2\sigma^2}  + \frac{1}{\sigma^2} |\la \BB\vv, \xixi \ra|
\;\; \overset{(\ref{TBound})}{<} \;\;
\frac{\varepsilon^2}{2\sigma^2} + \alpha
\;\; \overset{(\ref{TBound})}{=} \;\;
\frac{\varepsilon^2}{2\sigma^2}
+ \frac{\varepsilon}{\sigma}\sqrt{2 \ln \frac{2}{\delta}}
\;\; \leq \;\; q,
\ea
\eeq
where the last inequality is satisfied for any given $q \in (0, 1)$, and for
a corresponding sufficiently large $\sigma$. Namely, by the condition of the theorem, we have chosen:
$$
\ba{rcl}
\sigma & := & 
\frac{\varepsilon}{q} \max\Bigl\{1, 2 \sqrt{ 2\ln\frac{2}{\delta} } \Bigr\},
\ea
$$
which ensures~\eqref{DBoundQ}.

Now, to justify the unlearning bounds~\eqref{UnlearningBounds}, we consider the set
$$
\ba{rcl}
S & := & \Bigl\{ \xixi \; : \; D_{\vv}(\xixi) < q \Bigr\}.
\ea
$$
Notice that for $\xixi \in S$, we have
\beq \label{ProbBound}
\ba{rcl}
p(\xixi) & \leq & e^{q} \cdot p(\xixi + \vv),
\ea
\eeq
while by the previous observations, the measure of the complement is small:
\beq \label{ComplementSmall}
\ba{rcl}
\overline{S}
\;\; = \;\;
\Bigl\{ \xixi \; : \; D_{\vv}(\xixi) \geq q \Bigr\},
\qquad
\P(\overline{S}) \;\; \leq \;\; \delta.
\ea
\eeq
Hence,
$$
\ba{rcl}
\P(\yy_K \in Y)
& = &
\P(\xi \in Y - \xx_K)
\;\; = \;\;
\int\limits_{Y - \xx_K}
p(\xixi)d\xixi
\;\; = \;\;
\int\limits_{S \cap (Y - \xx_K)}
p(\xixi)d\xixi
\; + \;
\int\limits_{\overline{S} \cap (Y - \xx_K)}
p(\xixi)d\xixi \\
\\
& \overset{(\ref{ComplementSmall})}{\leq} &
\int\limits_{S \cap (Y - \xx_K)}
p(\xixi)d\xixi \; + \; \delta
\;\; \overset{(\ref{ProbBound})}{\leq} \;\;
e^{q} \cdot \int\limits_{S \cap (Y - \xx_K)}
p(\xixi + \vv)d\xixi \; + \; \delta \\
\\
& \leq & 
e^{q} \cdot \int\limits_{Y - \xx_K}
p(\xixi + \vv)d\xixi \; + \; \delta 
\;\; = \;\;
e^{q} \cdot \P\bigl(\xi \in Y - \xx_f^{\star} \bigr) \; + \; \delta 
\;\; = \;\;
e^{q} \cdot \P\bigl(\yy^{\star} \in Y\bigr) \; + \; \delta,
\ea
$$
which is the first inequality~\eqref{UnlearningBounds}. The proof of the second inequality in~\eqref{UnlearningBounds} is identical due to symmetry.

Finally, to ensure the optimization guarantee~\eqref{BoundSolution}, we observe that
$$
\ba{rcl}
\E\bigl[ \| \yy_K - \xx^{\star}_f \|^2  \bigr]
& = &
\E\bigl[ \| \vv + \xixi \|^2  \bigr]
\;\; = \;\;
\E\bigl[ \| \vv \|^2 + 2 \la \BB \vv, \xixi \ra + \| \xixi \|^2  \bigr] \\
\\
& = & 
\| \vv \|^2 + \E\bigl[ \| \xixi \|^2 \bigr]
\;\; \overset{(\ref{VBounded})}{\leq} \;\;
\varepsilon^2 + \E\bigl[ \| \xixi \|^2 \bigr]
\;\; \leq \;\; \varepsilon^2 + \sigma^2 d,
\ea
$$
where in the last bound we used the equivalent 
representation $\xixi = \sigma \BB^{-1/2} \zz$ 
with the standard normal vector $\zz \in \mathcal{N}(\vzero, \II_d)$,
which leads to the mean of the standard $\chi^2(d)$ distribution:
$$
\ba{rcl}
\E\bigl[ \| \xixi \|^2 \bigr]
& = & \sigma^2 \E\bigl[ \| \zz \|_2^2  \bigr]
\;\; = \;\; \sigma^2 d.
\ea
$$
It remains to use Jensen's inequality for concave function $\sqrt{\cdot}$, in order to obtain~\eqref{BoundSolution},
which completes the proof.
\qed
\end{document}